\documentclass[conference]{IEEEtran}

\usepackage{cite}
\ifCLASSINFOpdf
\else
\fi

\usepackage{xcolor}
\usepackage[cmex10]{amsmath}
\usepackage{amssymb}
\usepackage{booktabs}

\usepackage{color}
\usepackage{bm}
\usepackage{bbm}
\usepackage{caption, subcaption}
\usepackage{graphicx} 
\usepackage{algorithmic,algorithm}

\newcommand{\bx}{\bm{x}}
\newcommand{\by}{\bm{y}}

\newcommand{\BX}{\mathbf{X}}

\usepackage{multirow}

\begin{document}
%
\title{Node Copying for Protection Against Graph Neural Network Topology Attacks}

\author{\IEEEauthorblockN{Florence Regol, Soumyasundar Pal and Mark Coates}
\IEEEauthorblockA{Department of Electrical and Computer Engineering\\
McGill University, 3480 University Street, Montreal, Quebec, Canada H3A2A7\\
Email: \{florence.robert-regol, soumyasundar.pal\}@mail.mcgill.ca,   mark.coates@mcgill.ca}}

%


\maketitle

\begin{abstract}
Adversarial attacks can affect the performance of existing deep learning models. With the increased interest in graph based machine learning techniques, there have been investigations which suggest that these models are also vulnerable to attacks.
In particular, corruptions of the graph topology can degrade the performance of graph based learning algorithms severely. This is due to the fact that the prediction capability of these algorithms relies mostly on the similarity structure imposed by the graph connectivity. Therefore, detecting the location of the corruption and correcting the induced errors becomes crucial. 
There has been some recent work which tackles the detection problem, however these methods do not address the effect of the attack on the downstream learning task.
In this work, we propose an algorithm that uses node copying to mitigate the degradation in classification that is caused by adversarial attacks.
The proposed methodology is applied only after the model for the downstream task is trained and the added computation cost scales well for large graphs. Experimental results show the effectiveness of our approach for several real world datasets.
\end{abstract}
\begin{keywords}
Adversarial attacks, Graph convolutional networks, Semi-supervised learning
\end{keywords}

\IEEEpeerreviewmaketitle

\section{Introduction}

The application of deep learning models in real world systems has become increasingly prevalent, and as a result there has been an increased attention paid to their robustness and vulnerability to adversarial attack~\cite{carlini2017}. It has been demonstrated that many deep neural networks are susceptible to malicious attack and this has given rise to serious concerns regarding their reliability.

In many problem domains, including recommender systems, fraud detection, disease outcome and drug interaction prediction, there are structural relationships between data items. A graph is a natural mechanism for representing these relationships and this has led to the desire to translate the success of neural networks to the graph setting. An intense research effort has led to many models and algorithms~\cite{defferrard2016, kipf2017, hamilton2017b, monti2017, gilmer2017, velivckovic2018, battaglia2018,zhang2019}. It has been demonstrated that knowledge of the graph can be leveraged to compensate for having limited access to labelled data. Subsequently, there has been successful industrial application of these models~\cite{ying2018, geng2019spatiotemporal, liu2019geniepath}. This has raised concerns regarding the vulnerabilities of graph neural networks (GNNs) and researchers have commenced the development and investigation of attacks and defence mechanisms. Understanding the adversarial vulnerabilities of GNNs helps to expose the limitations of existing GNN models and can inspire better models and training strategies~\cite{deng2019, sun2019b}. 

Convolutional neural networks are usually subject to attacks that involve data manipulation to alter features. Graph neural networks can be targeted by similar attacks, but they are also subject to an alternative form of attack that involves alteration of the graph topology. In~\cite{zugner2018}, Z\"{u}gner et al.\ proposed Nettack, a method for constructing adversarial perturbations of graph data, which alters the graph topology and/or the node attributes in order to produce significant degradation in node classification performance. The experimental analysis in~\cite{zugner2018} suggests that attacks on the graph topology can have a more severe impact on classification performance compared to feature alteration. The attack in~\cite{zugner2018} strives to disrupt the classification of individual nodes in the graph; more recent work has targeted the deterioration of performance across the entire graph~\cite{zugner2019}. Other adversarial attacks on graphs have been proposed that highlight the vulnerability of GNNs for a wider range of inference tasks. In~\cite{dai2018}, Dai et al.\ show the efficacy of their proposed method on a real-world financial dataset where the classification task is to distinguish normal transactions from abnormal ones. This practical scenario gives a concrete example of how harmful such attacks can be and motivates the need for designing efficient countermeasures.

In response to the development of attacks on graph learning, there has been some preliminary research into detecting attacks. Zhang et al.~\cite{zhang2019b} propose an algorithm to detect which nodes have been subjected to an attack via modification of their edges. The procedure relies on the inconsistencies that the attack induces in the  classification outputs in the neighbourhood of an attacked node. Although the technique in~\cite{zhang2019b} offers a promising (albeit not foolproof) approach for detection of an attack, it does not provide a mechanism for rectifying the output of the learning algorithm.

In this paper, we focus on the next stage in the learning pipeline. We address the question of what to do after a detection procedure has notified us that there is a high probability that a node has been subjected to a topology attack. We introduce a copying procedure to partially recover the model accuracy of a graph convolutional neural network (GCN) for the corrupted nodes. The procedure involves copying the features of an attacked node to multiple similar locations in the graph and evaluating the output at these locations. The intuition is that when the features are moved to locations that correspond to its true class and have not been attacked, the GCN will return a correct classification.  
Through analysis of citation network datasets, we illustrate that this procedure can improve the classification accuracy by 10-15 percent for the attacked nodes.  

The paper is organized as follows. In Section II, we present background material, briefly reviewing graph convolutional neural networks. Section III provides more detail regarding the problem setting and Section IV presents our proposed recovery methodology. Section V describes the numerical experiments and presents and discusses the results. Section VI concludes the paper and suggests future research directions.

\section{Graph Convolutional Networks}\label{background}
For the scope of this paper, we address a downstream node classification task using a GCN proposed in~\cite{defferrard2016, kipf2016}.
In this setting, we are given a set of nodes $\mathcal{V}$ and edges $\mathcal{E} \subseteq \mathcal{V} \times \mathcal{V}$ that form a graph $\mathcal{G}_{obs} = (\mathcal{V},\mathcal{E})$. Node $i$ is associated with a feature vector $\bx_i$ and a label $\by_i$. 

In the semi-supervised setting, we have knowledge of labels only at a limited subset of nodes, $\mathcal{V}_{train}\subset \mathcal{V}$, and we aim to predict the labels at the nodes at the test set, $\mathcal{V}_{test}\subset \mathcal{V}$.  
The model uses information provided by the observed graph $\mathcal{G}_{obs}$, the complete feature matrix $\BX = [\bx_1, \bx_2, \dots, \bx_N]^T$ and the labels in the training set  $\mathbf{Y_{\mathcal{V}_{train}}} = \{\by_i: i \in \mathcal{V}_{train}\}$.

The layerwise propagation rule in simpler GCN architectures~\cite{defferrard2016, kipf2016} is based on a graph convolution operation and can be written as:
\begin{align}
\mathbf{H}^{(1)} &=\sigma(\mathbf{\hat{A}}_{\mathcal{G}}\mathbf{X}\mathbf{W}^{(0)}) \,,\\
\mathbf{H}^{(l+1)} &= \sigma(\mathbf{\hat{A}}_{\mathcal{G}}\mathbf{H}^{(l)}\mathbf{W}^{(l)})\,. 
\end{align}
Multiplication with the normalized adjacency operator $\mathbf{\hat{A}}_{\mathcal{G}}$ results in aggregation of the output features across node neighborhood at each layer.
$\mathbf{W}^{(l)}$ is a matrix of trainable weights at layer $l$ of the neural network and $\sigma$ denotes a pointwise non-linear activation function. $\mathbf{H}^{(l)}$ is the output representation from layer $l-1$.
In a node classification setting, using a $L$-layer network, the prediction is obtained by applying a softmax activation in the last layer and is written as $\widehat{\mathbf{Y}} = \mathbf{H}^{(L)}$. The weights of the neural network are learned via
backpropagation with the objective of minimizing the cross entropy loss between the training labels $\mathbf{Y}_{\mathcal{V}_{train}}$ and the network predictions
$\widehat{\mathbf{Y}}_{\mathcal{V}_{train}} =  \{\bm{\hat{y}}_i: i \in \mathcal{V}_{train}\}$ at the nodes in the training set.

\section{Problem Setting}

As stated in Section~\ref{background}, we address a semi-supervised node classification task. Based on the graph $\mathcal{G}_{obs}$, node features $\mathbf{X}$ and a small subset of known training labels $\mathbf{Y}_{\mathcal{V}_{train}}$, the goal is to infer the labels of the nodes in the test set, $ \mathcal{V}_{test} = \mathcal{V} \setminus \mathcal{V}_{train}$. However, a subset of the test nodes $ \mathcal{V}_{attacked} \subset \mathcal{V}_{test}$ are subjected to adversarial attack (details in Section~\ref{subsec:attack}), which modifies the graph. We consider a random poisoning attack~\cite{zugner2018} scenario where the attack precedes the model training. As a result, we only have access to the attacked graph $\mathcal{G}_{attacked}$. We assume that the attack only targets a small number of nodes compared to the size of the whole graph and it does not affect any nodes in $\mathcal{V}_{train}$.
For the scope of this work, we also assume that the identities of the nodes in $\mathcal{V}_{attacked}$ are known. In practice, this does not impose any serious restriction on the applicability of the proposed methodology since any reasonably accurate detection algorithm, such as the one proposed in~\cite{zhang2019b} can be employed to identify the nodes in $ \mathcal{V}_{attacked}$. If a node is incorrectly labelled as attacked, our proposed procedure in most cases does not modify the classification output. Our goal is to correct the possible classification errors for the nodes in $\mathcal{V}_{attacked}$ after the poisoning attack has occurred. 

\subsection{DICE Attack}
\label{subsec:attack}
Since the impressive performance of most graph based learning algorithms stems from the presence of edges between similar nodes, we consider an attack which aims to disrupt the similarity structure imposed by the graph connectivity. The DICE (Delete Internally Connect Externally) attack is a simple yet effective random attack.
It is parameterized by $0 < \beta \leq 1$, which dictates the severity of the degradation of the nodes in $\mathcal{V}_{attacked}$. We assume that the attacker has complete knowledge of the true labels of the nodes in $\mathcal{G}_{obs}$. For each attacked node $v$ with degree $d_{v}$, the attacker removes 
$\left \lceil{\beta d_{v}}\right \rceil$ of its existing edges at random and inserts new edges between node $v$ and $\left \lceil{\beta d_{v}}\right \rceil$ other nodes, sampled uniformly from the set of all nodes with different true labels from $v$. As a result, node $v$ has at most 
$\left \lfloor{(1-\beta)d_{v}}\right \rfloor$ neighbours with the same label after the attack. Since this attack does not perturb the degree of the target node, the degree distribution of the nodes in $\mathcal{V}_{attacked}$ remains unaltered.

\section{Methodology}

In our correction strategy, the classification of each node in $\mathcal{V}_{attacked}$ is performed in the following way. First we train a base GCN classifier using the small subset of labeled nodes $\mathbf{Y}_{\mathcal{V}_{train}}$ on $\mathcal{G}_{attacked}$ and store the obtained model.
Then we compute a lower dimension representation of the nodes of $\mathcal{G}_{attacked}$ using a node embedding algorithm. This procedure summarizes the information provided by the graph connectivity and the node features in the embedding. In our experiments, we use the Graph Variational Auto-Encoder (GVAE)~\cite{kipf2017} to obtain the embeddings but any other suitable techniques can also be employed. 

\begin{figure}[htbp]
\centering
\includegraphics[trim={0 0 0 1em}, scale=0.3, clip]{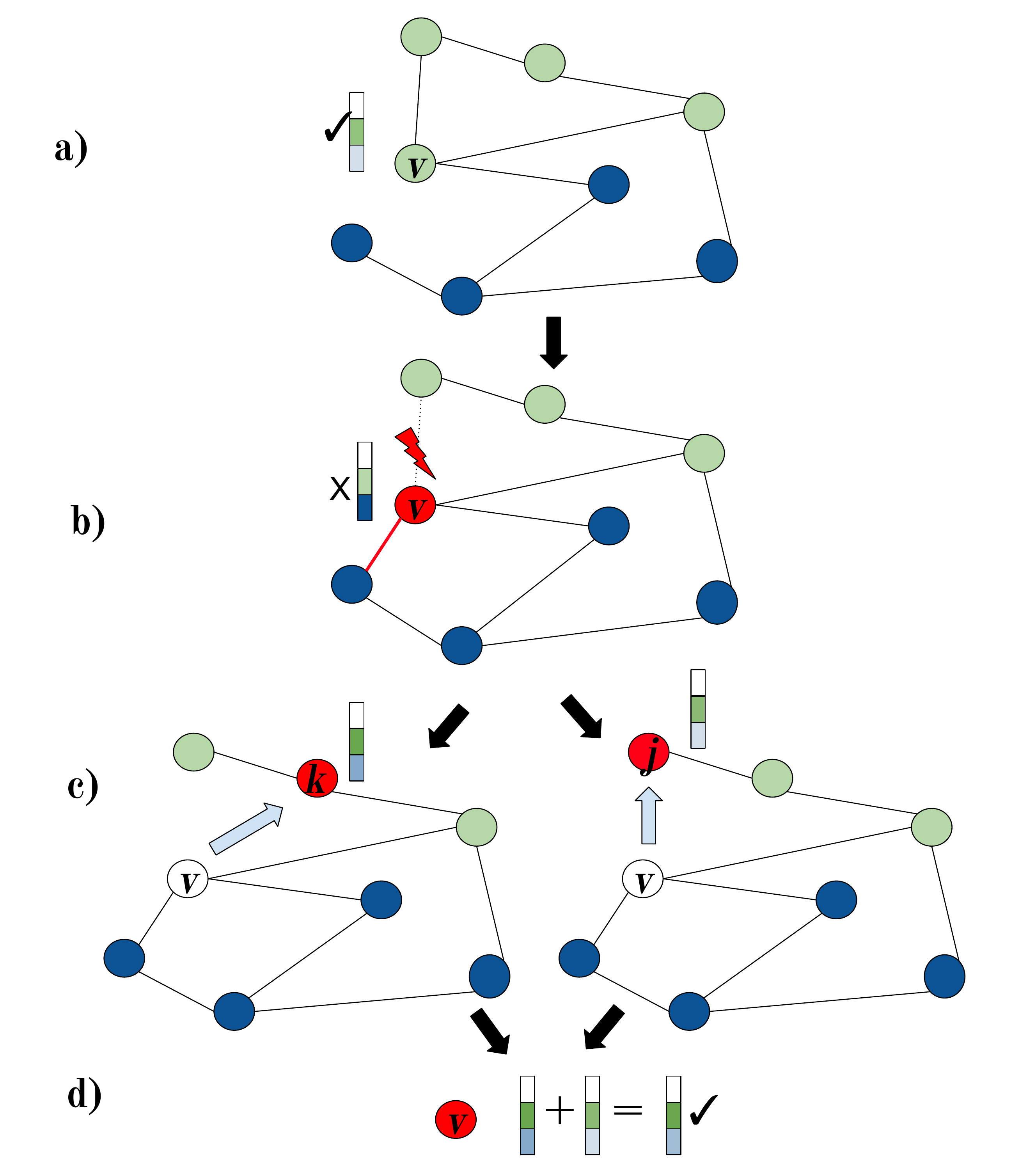}
\caption{Summary of the node copying procedure. \textbf{a)} In the absence of the attack, the softmax of node $v$ achieves the correct classification in $\mathcal{G}_{obs}$. \textbf{b)} Node $v$ is targeted by an attack and is now wrongly classified. \textbf{c)} The feature of node $v$ is copied to two new positions $k$ and $j$ and the softmax at those positions $\bm{\hat{y}}_{v \rightarrow k}$ and $\bm{\hat{y}}_{v \rightarrow j}$ are obtained. \textbf{d)} The error on node $v$ is corrected by computing the average of $\bm{\hat{y}}_{v \rightarrow k}$ and $\bm{\hat{y}}_{v \rightarrow j}$.} 
\label{fig:node_copying}
\end{figure}

We form a symmetric, pairwise distance matrix $D \in \mathbf{R}_{+}^{N \times N}$, whose $(i,j)$-th entry is defined as:
\begin{align}
    D_{i.j} = \|e_i - e_j\|_2 \,.\label{eq:dist}
\end{align}
Here $e_i$ is the embedding of node $i$ and $D_{i,j}$ is the distance between node $i$ and $j$. This distance matrix $D$ is subsequently used to select a set of similar nodes for each node in $\mathcal{V}_{attacked}$. For node $v$, we form the set $\mathcal{V}_{v}^{(p)}$ of the $p$ most similar nodes based on the $p$ lowest distances from node $v$ as follows:
\begin{align}
\mathcal{V}_{v}^{(p)} = \{ 1 \leq k \leq N \mid k \neq v, D_{i,k} = D_{i,(j)},  1 < j \leq p+1 \}\,,\label{eq:similar_set}
\end{align}
where $D_{i,(j)}$ is the $j$-th order statistic of $\{D_{i,k}\}_{k=1}^N$. Then for each node $k \in \mathcal{V}_{v}^{(p)}$, we copy the feature at node $v$ to  node $k$ (which is equivalent to copying the $v$-th row of $\BX$ to the $k$-th row) and compute the prediction $\bm{\hat{y}}_{v \rightarrow k}$ at node $k$ using the existing GCN model. The prediction for node $v$ using the proposed copying procedure is obtained by computing the average of $\{\bm{\hat{y}}_{v \rightarrow k}\}_{k \in \mathcal{V}_{v}^{(p)}}$. Figure~\ref{fig:node_copying} presents an overview of the complete procedure.
\vspace{-1em}

This correction procedure is successful if, on average, the true class of node $v$ is dominating in the set of softmax outcomes. The intuition is that some of the similar nodes included in $\mathcal{V}^{(p)}_v$ will have the same class as the true class $t$ of node $v$. We take the simplified view that, if uncorrupted, a node will often have the same class as most of its neighbors. If the node $v$ is copied at node $k$ in a ``wrong" neighborhood, meaning that node $v$ has a different class $w \neq t$ from node $k$ and its neighbors, then pulling the classification of node $v$ to the wrong class $w$ is usually harder and the $w$-th entry in the resulting softmax $\hat{\mathbf{y}}_{v \rightarrow k}$ is likely to be smaller.
 However when a node is placed in a ``good" neighborhood, the weighted average operation should reinforce the model confidence in the correct class $t$. So when we perform the average over softmax outputs at similar nodes, correct classification can be recovered.

A naive implementation of this method requires $p|\mathcal{V}_{attacked}|$ additional GCN evaluations after the training. This computational burden might be prohibitive for large graphs. However, we note that the prediction at any particular node from an $L$ layer GCN is influenced only by the $L$-hop neighbourhood of the node, which allows a much cheaper, localized computation of $\{\bm{\hat{y}}_{v \rightarrow k}\}$. The procedure is summarized in Algorithm~\ref{alg:copying}.

\begin{algorithm}[htbp]
\caption{Error correction using node copying}
\label{alg:copying}
\begin{algorithmic}[1]
\STATE {\bfseries Input:}  $\mathcal{G}_{attacked}$, $\BX$, $\mathbf{Y_{\mathcal{V}_{train}}}$, $\mathcal{V}_{attacked}$
\vspace{0.1cm}
\STATE {\bfseries Output:}  $\widehat{\mathbf{Y}}_{attacked}^{GCN}$, $\widehat{\mathbf{Y}}_{attacked}^{Copying}$
\vspace{0.1cm}
\STATE Train a GCN using $\mathcal{G}_{attacked}, \BX, $ $\mathbf{Y_{\mathcal{V}_{train}}}$ and compute $\widehat{\mathbf{Y}}_{attacked}^{GCN} = \{\bm{\hat{y}}_{v}\}_{v \in \mathcal{V}_{attacked}}$.

\STATE Train a GVAE to obtain node embeddings, $\{e_i\}_{i=1}^N$. Compute the pairwise distance matrix $D$ using eq.~\eqref{eq:dist}.

\FOR{$v \in \mathcal{V}_{attacked}$}
\STATE Form the set $\mathcal{V}_{v}^{(p)}$ using eq.~\eqref{eq:similar_set}
\FOR{$k \in \mathcal{V}_{v}^{(p)}$}
\STATE Copy the features of node $v$ in place of node $k$ and compute $\bm{\hat{y}}_{v \rightarrow k}$ using the existing GCN weights.
\ENDFOR
\STATE Compute $\bm{\hat{y}}_{v}^{Copying} = \frac{1}{p}\sum_{k \in \mathcal{V}_{v}^{(p)}} \bm{\hat{y}}_{v \rightarrow k}$
\ENDFOR
\STATE  Form $\widehat{\mathbf{Y}}_{attacked}^{Copying} = \{\bm{\hat{y}}_{v}^{Copying}\}_{v \in \mathcal{V}_{attacked}}$
\end{algorithmic}
\end{algorithm}

\section{Experiments}

We conduct experiments on three citation datasets: Pubmed, Citeseer and Cora~\cite{sen2008}. The prediction task is to classify the topics of research articles. Each document is represented as a node in a graph that is formed by adding an edge between any two articles if one of them cites the other. The features consist of a bag-of-words vectors extracted from the contents of the articles. Statistics of the datasets can be seen in Table~\ref{table:datasets}.

\begin{table}[b]
\centering
\caption{Datasets statistics}
\begin{tabular}{ccccl}
\hline
\textbf{Dataset} & \textbf{Nodes} & \textbf{Classes} & \textbf{Edges} & \textbf{Features} \\ \hline
\textbf{Cora} & 2,708 & 7 & 4,732 & 3,703 \\
\textbf{Citeseer} & 3,327 & 6 & 5,429 & 1,433 \\
\textbf{Pubmed} & 19,717 & 3 & 44,338 & 500 \\ \hline
\end{tabular}
\label{table:datasets}
\end{table}

The purpose of our attack correction algorithm is to retain the advantages derived from the model's ability to exploit knowledge of the graph topology. The information from the graph is more valuable when the amount of labelled data is severely limited, so we focus on this setting.

For each trial, $\mathcal{V}_{train}$ is formed by randomly sampling 10 or 20 nodes per class. Then an additional 50 nodes are sampled from the remaining set of nodes and these are targeted by the attack.
The rows in the adjacency matrix of $\mathcal{G}_{obs}$ of each node in $\mathcal{V}_{train}$ are iteratively corrupted following the DICE attack described previously. 
We consider the parameters $\beta = 0.5$ and $\beta = 0.75$ for the attack to test the robustness of the recovery procedure. The number of new positions for a node $p$ is set to 10 in our experiment. This parameter can be chosen more judiciously through cross-validation.
The GCN and GVAE hyperparameters are set to the values specified in~\cite{kipf2017} and~\cite{kipf2016}, respectively. These are obtained by optimizing the classification accuracy on a validation set of 500 nodes on the Cora dataset.

For each setting, we conduct 50 random trials, each of which corresponds to a random sampling of the training and attacked nodes and a random initialization of the GCN and the GVAE weights. We compare the accuracy on $V_{attacked}$ before and after copying. ``Before copying'' refers to the case where we collect the prediction for the attacked nodes from the GCN, which is trained on the attacked graph $\mathcal{G}_{attacked}$.
In addition, we report a graph agnostic baseline ``Neural Network'' on the $V_{attacked}$ set to explore whether it is better to ignore the graph altogether after an attack has been detected. 

We employ a Wilcoxon signed-rank test to evaluate the statistical significance of the results obtained. All such tests are performed by comparing with the ``Before copying" results. Results marked with an asterisk (*) indicate settings where the test failed to declare a significance at the $5\%$ level.
\vspace{-0.1cm}
\subsection{Ablation Studies}
\vspace{-0.1cm}
We perform two ablation studies to validate the relevance of the components of the proposed procedure.

\subsubsection{Majority voting}
To evaluate the utility of averaging the softmax outputs, we compare with a method when classification is obtained by majority voting.
In this method, instead of averaging the softmax outputs at the similar nodes, we make a global decision according to a majority vote among the labels obtained at each nodes. Ties are resolved by random selection. 

\subsubsection{No copying}
The goal of the second ablation experiment is to ensure that this method is not simply relying on the clustering capability of the chosen embedding technique. \\
The procedure is the same up to the point where we copy the node. Now, instead of copying the features of the attacked node, we directly take the GCN output of the nodes in $\mathcal{V}_{v}^{(p)}$ and repeat the two classification procedures: averaging softmax and majority voting.
\vspace{-0.15cm}

\subsection{Results}
\vspace{-0.15cm}
\begin{table}[htbp]
\caption{Average accuracy of the attacked nodes :  training with 10 labels per class, $\beta = 50\%$ }
\setlength{\tabcolsep}{5pt}
\centering
\begin{tabular}{|c|c|c|c|}
 \hline
\textbf{Dataset} & \textbf{Before Copying} & \textbf{\begin{tabular}[c]{@{}c@{}}Copying\\ Average Softmax\end{tabular}}  & \textbf{Neural Network} \\ \hline
Cora & 51.2$\pm$8.0 & 56.2$\pm$6.2 & 48.9$\pm$7.2\\
Citeseer & 42.2$\pm$7.7 & 50.3$\pm$7.6 & 39.4$\pm$10.0 \\
Pubmed & 51.7$\pm$6.7 & 63.3$\pm$6.0 &  65.8$\pm$6.6\\ \hline
\end{tabular}

\label{table:10}
\end{table}

\vspace{-0.3cm}
\begin{table}[h]
\caption{Ablation study for majority voting : average accuracy of the attacked nodes for Cora. }
\setlength{\tabcolsep}{5pt}
\centering
\begin{tabular}{|c|c|cc|c|}
\hline
\textbf{\begin{tabular}[c]{@{}c@{}}Labels\\ per class\end{tabular}} & \textbf{$\beta$} & \textbf{\begin{tabular}[c]{@{}c@{}}Before\\ Copying\end{tabular}} & \textbf{\begin{tabular}[c]{@{}c@{}}Copying\\ Majority Voting\end{tabular}} & \textbf{\begin{tabular}[c]{@{}c@{}}Copying\\ Average Softmax\end{tabular}} \\ \hline
\multirow{2}{*}{10} & 50\% & 51.2$\pm$8.0 & 55.7$\pm$6.7 & 56.2$\pm$6.2 \\
 & 75\% & 38.8$\pm$6.4 & 38.3$\pm$8.4 & 39.3$\pm$8.0 \\ \hline
\multirow{2}{*}{20} & 50\% & 58.2$\pm$6.9 & 58.0$\pm$7.5* & 59.1$\pm$7.5* \\
 & 75\% & 32.4$\pm$6.4 & 38.3$\pm$7.3 & 39.1$\pm$7.3 \\ \hline
\end{tabular}
\label{table:maj_cora}
\end{table}

\vspace{-0.2cm}
\begin{table}[H]
\caption{Ablation study for majority voting : average accuracy of the attacked nodes for Citeseer.}
\setlength{\tabcolsep}{5pt}
\centering
\begin{tabular}{|c|c|cc|c|}
\hline
\textbf{\begin{tabular}[c]{@{}c@{}}Labels\\ per class\end{tabular}} & \textbf{$\beta$} & \textbf{\begin{tabular}[c]{@{}c@{}}Before\\ Copying\end{tabular}} & \textbf{\begin{tabular}[c]{@{}c@{}}Copying\\ Majority Voting\end{tabular}} & \textbf{\begin{tabular}[c]{@{}c@{}}Copying\\ Average Softmax\end{tabular}} \\ \hline
\multirow{2}{*}{10} & 50\% & 42.2$\pm$7.7 & 50.0$\pm$7.6 & 50.3$\pm$7.6 \\
 & 75\% & 29.0$\pm$6.9 & 40.4$\pm$6.1   &  40.2$\pm$5.9\\ \hline
\multirow{2}{*}{20} & 50\% & 48.7$\pm$5.9 & 52.7$\pm$6.3 & 53.1$\pm$7.0 \\
 & 75\% & 31.1$\pm$7.0 & 43.8$\pm$7.2 & 44.2$\pm$6.3\\ \hline
\end{tabular}
\label{table:maj_citeseer}
\end{table}

\begin{table}[htbp]
\caption{Ablation study no copying : Average accuracy of the attacked nodes for Cora}
\setlength{\tabcolsep}{2pt}
\centering
\begin{tabular}{|c|c|cc|c|}
\hline
\textbf{\begin{tabular}[c]{@{}c@{}}Labels\\ per class\end{tabular}} & \textbf{$\beta$} & \textbf{\begin{tabular}[c]{@{}c@{}}No Copying\\ Average Softmax\end{tabular}} & \textbf{\begin{tabular}[c]{@{}c@{}}No Copying\\ Majority Voting\end{tabular}} & \textbf{\begin{tabular}[c]{@{}c@{}}Copying\\ Average Softmax\end{tabular}} \\ \hline
\multirow{2}{*}{10} & 50\% & 44.2$\pm$6.5 & 43.4$\pm$6.4 & 56.2$\pm$6.2 \\
 & 75\% & 21.2$\pm$6.3 & 21.2$\pm$5.8 & 39.3$\pm$8.0 \\ \hline
\multirow{2}{*}{20} & 50\% & 45.2$\pm$8.6 & 44.7$\pm$7.8 & 59.1$\pm$7.5* \\
 & 75\% & 20.1$\pm$5.1 & 20.2$\pm$5.2 & 39.1$\pm$7.3\\ \hline
\end{tabular}
\label{table:non_cora}
\end{table}

\begin{table}[htbp]
\setlength{\tabcolsep}{2pt}
\caption{Ablation study no copying : Average accuracy of the attacked nodes for Citeseer}
\setlength{\tabcolsep}{2pt}
\centering
\begin{tabular}{|c|c|cc|c|}
\hline
\textbf{\begin{tabular}[c]{@{}c@{}}Labels\\ per class\end{tabular}} & \textbf{$\beta$} & \textbf{\begin{tabular}[c]{@{}c@{}}No Copying\\ Average Softmax\end{tabular}} & \textbf{\begin{tabular}[c]{@{}c@{}}No Copying\\ Majority Voting\end{tabular}} & \textbf{\begin{tabular}[c]{@{}c@{}}Copying\\ Average Softmax\end{tabular}} \\ \hline
\multirow{2}{*}{10} & 50\% & 35.6$\pm$6.4 & 35.6$\pm$6.4 & 50.3$\pm$7.6 \\
 & 75\% &21.3$\pm$5.6  &21.1$\pm$6.3  & 40.2$\pm$5.9 \\ \hline
\multirow{2}{*}{20} & 50\% & 36.8$\pm$5.5 & 37.7$\pm$5.4 & 53.1$\pm$7.0 \\
 & 75\% & 22.5$\pm$5.8 & 22.5$\pm$5.6 & 44.2$\pm$6.3\\ \hline
\end{tabular}
\label{table:non_citeseer}
\end{table}

\subsection{Discussion}
From Tables~\ref{table:10},~\ref{table:maj_cora} and~\ref{table:maj_citeseer}, we observe that the proposed algorithm offers significant improvement across all datasets from the ``Before Copying'' baseline at the attacked nodes.
In Table~\ref{table:10}, the relative advantage is apparent as the method is able to improve accuracy by between $5\%$ and $11.6\%$ while outperforming the neural network in most cases. This illustrates the capability of the method of being able to leverage the graph for classification in a situation where labeled data is scarce and not sufficient to train a competitive graph agnostic method.

The ablation studies also confirm that averaging over the softmax performs better than majority voting for almost all experimental settings, but the performance difference is small (Tables~\ref{table:maj_cora} and~\ref{table:maj_citeseer}). For the ``No Copying" ablation experiment, the results in Tables~\ref{table:non_cora} and~\ref{table:non_citeseer} show that the proposed copying mechanism offers significant improvement compared to simply using the outputs at the similar nodes. In some cases, ``No Copying" is even worse than the performance of ``Before Copying'', for both softmax averaging and majority voting.

\section{Conclusion}
In this paper, we have proposed a recovery algorithm which shows promising results in classifying nodes that have been subjected to a targeted topology attack. The post-attack classification step adds negligible overhead to overall training procedure. We have conducted experiments and ablation studies to highlight the relative importances of different components of the methodology.
This work can be further extended by combining the method with an attack detection technique; this will eliminate the assumption that we know the nodes that have been attacked. This scenario is a more realistic setting that we could expect to encounter in practice. In addition, another important research direction is to examine how the depends on the choice of embedding technique and graph-based classifier.




\bibliographystyle{IEEEtran}
\bibliography{reference}




\end{document}